\documentclass[letterpaper, 10 pt, conference]{ieeeconf}  

\IEEEoverridecommandlockouts                              

\usepackage{color}
\usepackage{graphicx}
\usepackage{tabularx}
\usepackage{crop}
\usepackage{multirow} 
\usepackage{caption}
\usepackage{amsmath}
\usepackage{amssymb}
\usepackage{xspace}
\usepackage{subfig}
\usepackage{booktabs}
\usepackage{dsfont}
\DeclareMathOperator*{\argmax}{arg\,max}
\DeclareMathOperator*{\argmin}{arg\,min}
\usepackage[export]{adjustbox}

\newcolumntype{Y}{>{\centering\arraybackslash}X}

\def\eg{\textit{e.g.}\xspace}
\def\ie{\textit{i.e.}\xspace}

\title{\LARGE \bf Knowledge is Never Enough: \\ Towards Web Aided Deep Open World Recognition 
}

\author{Massimiliano Mancini$^{1,2}$, Hakan Karaoguz$^{3}$, Elisa Ricci$^{2,4}$, Patric Jensfelt$^{3}$, Barbara Caputo$^{5}$
\thanks{This work was partially supported by EU Centauro Project, The Swedish Foundation for Strategic Research (SSF) FACT project (H.K., P. J.) and the ERC project RoboExNovo (M. M., B.C.).}
\thanks{$^{1}$M. Mancini is with Sapienza University of Rome, Rome, Italy.{\tt\small mancini@diag.uniroma1.it}}%
\thanks{$^{2}$M. Mancini and E. Ricci are with Fondazione Bruno Kessler, Trento, Italy. {\tt\small eliricci@fbk.eu}}%
\thanks{$^{3}$H. Karaoguz and P. Jensfelt are with KTH Royal Institute of Technology, Stockholm, Sweden.  {\tt\small \{hkarao,patric\}@kth.se}}
\thanks{$^{4}$E. Ricci is with University of Trento, Trento, Italy.}
\thanks{$^{5}$B. Caputo is with Italian Institute of Technology, Milan, Italy.}
}

\begin{document}

\begin{titlepage}
\null
\vfill
\renewcommand{\fboxsep}{10pt}
\fbox{\Large\begin{minipage}{\columnwidth}
\textbf{Disclaimer:}

This work has been accepted for publication in the Proceedings of the 2019 IEEE International Conference on Robotics and Automation\vspace{4pt}
\newline
link:   https://www.icra2019.org/
\newline
\newline
If you want to cite this article, please use:\newline
\emph{\large @inProceedings\{mancini2019knowledge,\newline
	\hspace*{1cm} author = \{Mancini, Massimilano and Karaoguz, Hakan and Ricci, Elisa and Jensfelt, Patric \newline\hspace*{3cm} and Caputo, Barbara\},\newline\hspace*{1cm} 
  	title  = \{Knowledge is Never Enough: Towards Web Aided Deep Open World Recognition\},\newline\hspace*{1cm} 
  	booktitle = \{IEEE International Conference on Robotics and Automation (ICRA)\},\newline\hspace*{1cm} 
  	year      = \{2019\},\newline\hspace*{1cm} 
  	month     = \{May\}\newline
\}}
\newline
\newline
\textbf{Copyright:} 
\newline
\copyright~2019 IEEE. Personal use of this material is permitted. Permission from IEEE must be obtained for all other uses,  in  any  current  or  future  media,  including  reprinting/  republishing  this  material  for  advertising  or promotional purposes, creating new collective works, for resale or redistribution to servers or lists, or reuse of any copyrighted component of this work in other works.
\newline
\end{minipage}}
\vfill
\clearpage
\end{titlepage}
\setlength{\abovedisplayskip}{4pt}
\setlength{\belowdisplayskip}{4pt}

\maketitle
\thispagestyle{empty}
\pagestyle{empty}

\begin{abstract}
While today's robots are able to perform sophisticated tasks, they can only act on objects they have been
trained to recognize. This is a severe limitation: any robot will inevitably see new objects in unconstrained settings, and thus will always have visual knowledge gaps.  
However, standard visual modules are usually built on a limited set of classes and are based on the strong prior that an object must belong to one of those classes. 
Identifying whether an instance does not belong to the set of known categories (\ie open set recognition), only partially tackles this problem, as a truly autonomous agent should be able not only to detect what it does not know, but also to extend dynamically its knowledge about the world. We contribute to this challenge with a deep learning architecture that can dynamically update its known classes in an end-to-end fashion. The proposed deep network, based on a deep extension of a non-parametric model, detects whether a perceived object belongs to the set of categories known by the system and learns it without the need to retrain the whole system from scratch. Annotated images about the new category can be provided by an 'oracle' (i.e. human supervision), or by autonomous mining of the Web. Experiments on two different databases and on a robot platform demonstrate the promise of our approach.

\end{abstract}

\section{INTRODUCTION}
\label{sec:intro}

\vspace{4pt}
For robots to perform intelligent, autonomous behaviors, it is crucial that they understand what they see. The applications requiring visual abilities are countless: from
self-driving cars to detecting and handling objects for service robots in homes, from kitting in industrial workshops, to robots filling shelves and shopping baskets in supermarkets, etc,  they all  imply interacting with a wide variety of objects, requiring in turn a deep 
understanding of what
these objects look like, their visual properties and associated functionalities. Still, the best vision systems we have today are not yet up to the needs of artificial autonomous systems in the wild.
There are examples of robots performing complex tasks such as loading a dishwasher \cite{asfouretal08} or flipping pancakes \cite{beetzetal2011}. However, the visual knowledge about the objects involved in these tasks is manually encoded within the robots control programs
or knowledge bases, limiting them to operate on the objects they have been programmed to understand. More in general, the current mainstream approach to visual recognition, based on convolutional neural networks \cite{alexnet,vgg16}, makes the so called \emph{closed world assumption}, i.e. it assumes that the number and type of objects a robot will encounter in its activities is fixed and known a priori.
Hence, the big challenge is to make these visual algorithms robust to illumination, scale and categorical variations as well as clutter and occlusions. 

\begin{figure}[tb]
  \centering
  \includegraphics[width=1.0\columnwidth]{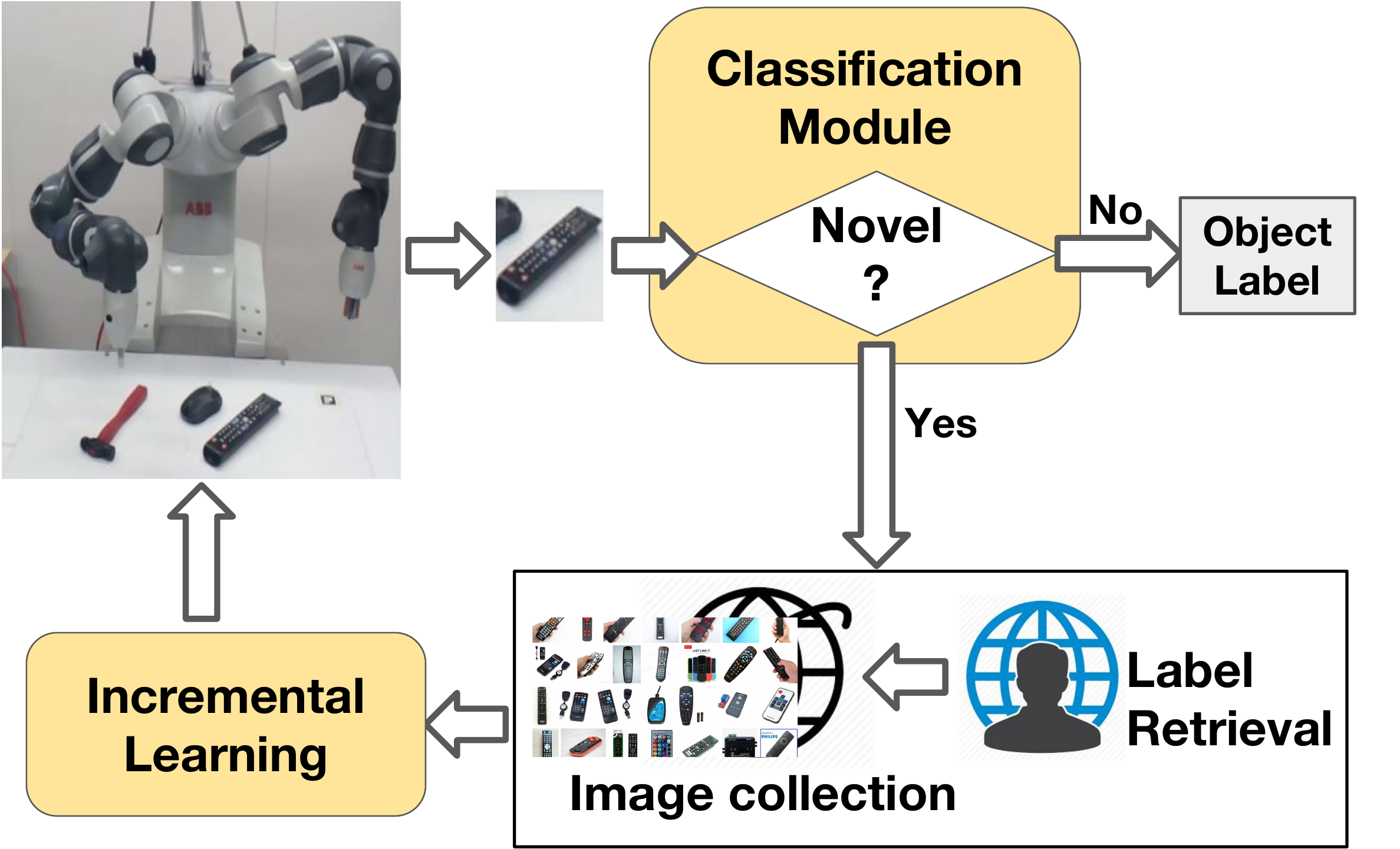}
 \caption{{Overview of the open world recognition task within a robotic platform. Given an image of an object, a classification algorithm assigns to it a class label. If the object is recognized as novel, the object label and relative are obtained through external resource (\eg a human and/or the Web). Finally, the images are used to incrementally updated the knowledge base of the robot.} 
 }
  \vspace{-20pt}
  \label{fig:teaser}
 \end{figure}

While these are crucial issues in robot vision, solving them is not enough. 
Any robot, regardless of how much knowledge has been manually encoded into it, will
inevitably see novel objects.
This calls for robots able to 
know what they know and what they do not know, and able to learn how to recognize new
objects by themselves. {For instance, in Fig.~\ref{fig:teaser}, a 2-arm manipulator robot detects a novel object on its workspace. It then obtains the object label and images of the same object through external resources (\eg a human collaborator or by mining the Web). Finally, the robot incrementally learns the novel object category and begins to detect the novel object correctly.}

  
  
The first contribution of this paper is a deep visual recognition algorithm that moves beyond the closed world assumption. Our work falls into the Open World Recognition (OWR) framework introduced by \cite{bendale2015towards},  where
the task is to learn a model able to classify images if they belong to the categories of the training set, to spot samples  corresponding to unknown classes, and on the basis of such unknown class detections  
update the model in order to progressively include the novel categories. We build on recent work by Guerriero et al \cite{guerriero2018deep} and present the first deep open world recognition architecture. Our approach couples the flexibility of non-parametric classification methods, necessary to add incrementally new classes over time and able to estimate a probability score for each known class supporting the detection of new classes (Nearest Non Outlier, NNO \cite{bendale2015towards}), with the powerful intermediate representations learned by deep networks. We enable end-to-end training of the architecture through an online approximate estimate and update function for the mean prototype representing each known class and for the threshold allowing to detect novel classes in a life-long learning fashion.   
We show experimentally that our algorithm outperforms previous OWR methods in terms of its ability to (a) detect novel classes and (b) add new classes to the set of known ones.

A key issue when attempting to overcome knowledge gaps is
how to get training data for the new classes. The OWR framework assumes the existence of an 'oracle', providing annotated images for each new class. In a robotic scenario, this has often translated into having a human in the loop, with the robot asking for images and labels. This scenario somehow limits the autonomy of a robot system, that without the presence of a teacher would find itself stuck when  detecting a new object. We propose here to augment the open ended learning abilities of robots by mining the Web for weakly annotated images of the new detected objects, mimicking the human ability to learn not only from situated experiences, but also from visual knowledge externalized on artifacts like drawings, or indeed Web resources. Hence, the second contribution of this paper is a simple 
protocol for mining the Web starting from images of objects classified as unknown, able to provide labeled data for initializing the learning of new classes on the fly.  
Experiments show that although the Web images do not provide the same informative content one would get from a human oracle, results are very promising. 
Further experiments
on a mobile robot platform demonstrate the power of our approach, that to our knowledge defines the new state-of-the-art in OWR while opening new perspectives in webly 
supervised learning for autonomous systems.
\vspace{-3pt}


\subsection{Related work}
\vspace{-5pt}
Convolutional Networks have greatly advanced the visual abilities of intelligent autonomous systems in the last years \cite{Behnke,eitel2015multimodal,carlucci2018deco,mancini2018kitting}. Still, in a similar fashion with what happened with shallow learning methods, the vast majority of algorithms assume that the entire training set is available beforehand (batch training), without the possibility to add knowledge about the known classes, or about new classes, in a life-long learning fashion.
Most of previous work in the robot vision literature used statistical learning tools, dealing with continuous learning of known classes for semantic spatial models \cite{pronobis2010ivc} or incremental class learning of object models \cite{pasquale2016object,camoriano2017incremental}. Alternative shallow approaches explored the suitability of unsupervised methods for incremental class learning of object categories \cite{ott2012unsupervised}. Only very recently the problem has been cast into the deep learning framework \cite{valipour2017incremental}. The visual learning community has investigated in the last few years the problem of life-long, open ended learning \cite{de2016online,rebuffi2017icarl}, with approaches tackling a unifying framework for novelty detection and incremental class learning (Open world Recognition, OWR \cite{bendale2015towards,he2016deep,guerriero2018deep}). While these efforts are principled and hold promise, to the best of our knowledge their effectiveness has never been tested within the robot vision scenario. 
Moreover, current deep approaches in this thread address mainly the incremental class learning problem, rather than the whole OWR challenge.  
\label{sec:related-work}

\section{Deep Open World Recognition}
\label{sec:method}
In this section we describe our proposed framework for OWR. Our framework contains two core components. The first is an OWR model, named Deep Nearest Non-Outlier (DeepNNO), which merges the NNO classification algorithm with the power of deep representations, within an end-to-end trainable deep architecture (section~\ref{sec:deepnno}). The second is a semi-automatic pipeline based on Web queries (section~\ref{sec:web-download}) which replaces the presence of an oracle giving to the algorithm both labels and data relative to novel classes, as assumed by previous OWR approaches~\cite{bendale2015towards}. 
In the following, we start by formalizing the OWR problem.

\vspace{-4pt}
\subsection{Problem Definition}
\vspace{-5pt}
As stated in Section \ref{sec:intro}, the aim of Open World Recognition is to learn a model able to: (i) correctly classify samples of known classes, 
for which the model has been trained on; (ii) detect instances belonging to novel classes, unseen during training; (iii) incrementally include novel
classes within the set of known ones, extending the capabilities of the classifier.

Formally, let us denote as
$\mathcal{T}_t=\{(x_i,k_i)\}$ the training set, where $x_i$ refers to the $i$-{th} image and $k_i$ its semantic label. 
In OWR the training samples are provided in an incremental fashion, \ie $\mathcal{T}_t$ denotes the training set at time $t$ with cardinality $M_t=|\mathcal{T}_t|$. Furthermore $k_i \in \mathcal{K}_t$ and the set of \emph{known categories} $\mathcal{K}_t$ is dynamically updated. Specifically, this set is initialized considering 
labels associated to the first $N$ semantic categories, \textit{i.e.}  $\mathcal{K}_0=\{1,\dots,N\}$, and it is progressively extended when novel classes are discovered. In summary, we denote the initial set of known classes as $\mathcal{K}_0$ and the set of known classes after $t$ incremental steps as $\mathcal{K}_t$, with $\mathcal{K}_{t}\subset\mathcal{K}_{t+1}$. Similarly let $\mathcal{T}_0$ denote the training set exploited to learn the initial classification model $f$ which is then updated with $\mathcal{T}_t$.

In OWR at each time step our goal is twofold. First, we want to detect if a new sample $x_n$ is an instance of a novel class, \ie if the associated label $k_n \in \mathcal{U}_t$, where $\mathcal{U}_t$ is the set of \textit{unknown categories} after $t$ incremental steps. 
{Second, 
we want to incrementally update our classification model $f$ such that $\mathcal{K}_{t+1}=\mathcal{K}_{t}\cup \{k_n\}$. 
In the following we describe the proposed technique to fulfill this goal. 
Before 
detailing our approach, we review the Nearest Class Mean (NCM) \cite{mensink2012metric} algorithm and previous works \cite{bendale2015towards} extending NCM to tackle
the OWR problem.

\vspace{-4pt}
\subsection{From Nearest Class Mean to Nearest Non-Outlier}
\vspace{-4pt}
The NCM algorithm~\cite{mensink2012metric} is a non-parametric classification algorithm which operates in traditional closed world setting (\ie no incremental learning and novel category discovery process take place). NCM computes a mean feature vector for each semantic category and assigns an input image to a class according to the distance between the image and the mean vectors in the feature space. 
Formally, let us denote with $\phi(\cdot)$ a function that, given an image $x$, extracts a feature vector in a $m$-dimensional space. 
The NCM algorithm assigns a class label $k^*$ to an input image computing:
\begin{equation}
\label{eq:ncm-prediction}
k^*=\argmin_{k \in \mathcal{K}}d_k(x)=\argmin_{k \in \mathcal{K}} ||\phi(x) - \mu_k ||
\end{equation}
where $\mu_k \in \Re^m$ is the mean feature vector relative to class $k\in \mathcal{K}$ in the training set, and $d_k(x)$ is the distance between 
the mean vector and the features extracted from the input sample. 
In \cite{mensink2012metric} the authors proposed to improve the performance of the traditional NCM by including a metric learning procedure. 
In particular, let us suppose to have a matrix $W\in \Re^{m\times d}$ which projects the feature vectors into a $d$-dimensional space. In \cite{mensink2012metric}, Mensink \textit{et al.} proposed to optimize $W$ by minimizing the following loss:
\begin{equation}
\label{eq:loss}
\mathcal{L}=-\frac{1}{M}\sum_{(x_i,k_i)\in \mathcal{T}} \log p_{k_i}(x_i)
\end{equation}
where $p_k(x)= e^{-\frac{1}{2} d_k^W(x)}$, $d_k^W(x) = ||W^T \phi(x) - W^T \mu_k ||$.


While NCM is 
an effective algorithm for addressing visual recognition tasks, it only works under a closed world assumption. To break this assumption, 
Bendale \textit{et al.} \cite{bendale2015towards} extended NCM to OWR settings, presenting the Nearest Non-Outlier algorithm (NNO). 
In NNO for each sample the estimated class distances $d_k^W(x), \ \forall k \in \mathcal{K}_t$, are used to compute a set of class-specific scores defined as: 
\begin{equation}
\label{eq:nno-score}
s_k(x)=\eta_\tau \left(1-\frac{d_k^W(x)}{\tau}\right)
\end{equation}
where $\tau$ is a user-defined parameter and $\eta_\tau$ is a normalization factor depending on $\tau$. These scores are interpreted as probabilities after applying a clamping function $p_k(x)=\max(0, s_k(x))$. 
where a clamping function is introduced as $s_k(x)$ could be negative. 
These probabilities are used to detect if a sample $x$ belongs to a novel class, by defining the following prediction function:
\begin{equation}
\label{eq:nno-prediction}
k^*=\begin{cases}
      u \in \mathcal{U}_t & \text{if}\ 
      p_k(x)=0 \ \ \ \forall k \in \mathcal{K}_t\\
      \argmax_{k\in\mathcal{K}_t} p_k(x) & \text{otherwise}
    \end{cases}
\end{equation}
where a sample belongs to the class $u$ if it is \textit{rejected} (\ie $p_k(x)=0$) by all known classes $k\in\mathcal{K}_t$.
In \cite{bendale2015towards} it is assumed that every time a sample is detected as an instance of a novel class, \ie $k^*=u$, an oracle provides the correct label and 
other images for the same class. These images are used to compute the mean feature vector for the novel class
$\mu_u$ which is then included in the classification model. 
In \cite{bendale2015towards} the matrix $W$ is estimated during training minimizing a loss as in Eqn. (\ref{eq:loss}) and then kept fixed during deployment. The parameter $\tau$ is 
obtained through cross-validation. 
 
Previous works ~\cite{bendale2015towards,de2016online} showed how NNO is an effective approach for OWR. However, this method is based on a shallow classification model. Here we demonstrate that the performances of NNO can be significantly boosted through deep architectures. We further show that, by automatically leveraging images from the Web, it is possible to 
build a 
practical OWR systems for visual recognition in robotics platforms, with very limited need of human assistance. 

\vspace{-8pt}
\subsection{Proposed Method}
\vspace{-4pt}
\subsubsection{DeepNNO}

\label{sec:deepnno}
The \textit{Deep Nearest Non-Outlier} algorithm is obtained from NNO with the following modifications: (i)
the feature extractor function is replaced with deep representations derived from neural network layers; (ii)
an online update strategy is adopted for the mean vectors $\mu_k$; 
(iii) an appropriate loss is optimized 
using stochastic gradient descent (SGD) methods in order to 
compute the feature representations and the associated class specific means.


Inspired by the recent work \cite{guerriero2018deep},
we 
replace the feature extractor function $\phi(\cdot)$ with deep representations derived from a neural network $\phi_\Theta(x)$
and define the class-specific probability scores as follows:
\begin{equation}
\label{eq:deep-nno-prob}
p_k(x)=\exp\left(-\frac{1}{2}||\phi_\Theta(x) - \mu_k ||\right) .
\end{equation}
Note that, differently from \cite{bendale2015towards}, we do not consider explicitly the matrix $W$ since this is replaced by the network parameters $\Theta$. 
Furthermore, we avoid to use a clamping function as this could hamper the gradient flow within the network. 
This formulation is similar to the NNO version proposed in \cite{de2016online} which have been showed to be more effective than that in \cite{bendale2015towards} for online scenarios. 

In OWR the classification model must be updated as new samples arrive. In DeepNNO this translates into incrementally updating the feature representations $\phi_\Theta(x)$ and defining an appropriate strategy for updating the class mean vectors. 
Given a mini-batch of samples $\mathcal{B}=\{(x_1,k_1),\dots, (x_b,k_b)\}$, we compute the mean vectors through:
\begin{equation}
\label{eq:deep-nno-means}
\mu_k^{t+1}=\frac{n_k\cdot\mu_k^{t}+n_{k,\mathcal{B}}\cdot\mu_k^{\mathcal{B}}}{n_k+n_{k,\mathcal{B}}}
\end{equation}
where $n_{k}$ represents the number of samples belonging to class $k$ seen by the network until the current update step $t$, $n_{k,\mathcal{B}}$ represents the number of samples belonging to class $k$ in the current batch and $\mu_k^{\mathcal{B}}$ represents the current mini-batch mean vector
relative to the features of class $k$.

Given the class-probability scores in DeepNNO we define the following prediction function:
\begin{equation}
\label{eq:nno-predictiondeep}
k^*=\begin{cases}
      u \in \mathcal{U}_t & \text{if}\ p_k(x) \leq \theta \ \ \ \forall k \in \mathcal{K}_t\\
      \argmax_{k\in\mathcal{K}_t} p_k(x) & \text{otherwise}
    \end{cases}
\end{equation}
where $\theta$ is a threshold which, similarly to the parameter $\tau$ in Eqn.(\ref{eq:nno-score}), regulates the number of samples that are assigned to a new class. While in \cite{bendale2015towards} $\tau$ is a user defined parameter which is kept fixed, in this paper we argue that a better strategy is to dynamically update $\theta$ since the feature extractor function and the mean vectors change during training. {Intuitively, 
while training the deep network, an estimate of $\theta$ can be obtained by looking at the probability score given to the ground truth class. If the score is higher than the threshold, the value of $\theta$ can be increased. Oppositely, the value of the threshold is decreased if the prediction is rejected. Specifically, given a mini-batch $\mathcal{B}$ we update $\theta$ as follows:
\begin{equation}
\label{eq:deep-nno-tau-update}
\theta^{t+1}=\frac{1}{t+1}\left(t\cdot\theta^t+\frac{1}{C_\mathcal{B}}\sum_{k\in\mathcal{K}_t}\bar{p}_{k,\mathcal{B}}\right)
\end{equation}
where 
$C_\mathcal{B}$ is the number of classes in $\mathcal{K}_t$ represented by at least one sample in $\mathcal{B}$ and $\bar{p}_{k,\mathcal{B}}$ is the weighted average probability score of instances of class $k$ within the batch. Formally we consider:
\vspace{-8pt}
\begin{equation}
\label{eq:deep-nno-tau-batch-avg}
\bar{p}_{k,\mathcal{B}}=\frac{1}{\eta_{\mathcal{B},k}}\sum_{i=1}^{\text{b}}w_{k,i}\cdot p_{k}(x_i)
\vspace{-4pt}
\end{equation}
where $\eta_{\mathcal{B},k}=\sum_{i=1}^\text{b} w_{k,i}$ is a normalization factor and:
\begin{equation}
\label{eq:nno-ws}
w_{k,i}=\begin{cases}
     w^+ & \text{if}\ k_i=k \wedge p_k(x_i)>\theta \\
     w^- & \text{if}\ k_i=k \wedge p_k(x_i)\leq \theta \\
      0 & \text{otherwise}
    \end{cases}
\end{equation}
where $w^-$ and $w^+$ are scalar parameters which allow to assign different importance to samples for which the scores given to the ground truth class are respectively rejected or not by the current threshold $\theta$. 

}

To train the network, we employ standard SGD optimization, minimizing the binary cross entropy loss over the training set:
\begin{equation}
\vspace{-8pt}
 \mathcal{L}=\frac{1}{|\mathcal{T}_t|}\sum_i\ell^{\text{cl}}(x_i,k_i) 
 \vspace{-5pt}
\end{equation}
where:
\begin{equation}
 \ell^{\text{cl}}(x_i,k_i) =  \log p_{k_i}(x_i) + \sum_{k \in \mathcal{K}_t} \mathds{1}_{k\neq k_i} \log \left(1 - p_k(x_i)\right)
\end{equation}
After computing the loss, we use standard backpropagation to update the network parameters $\Theta$. After updating $\Theta$, we use the samples of the current batch to update both the class mean estimates $\mu_k$ and the threshold $\theta$, 
using Eqn.\eqref{eq:deep-nno-means} and Eqn.\eqref{eq:deep-nno-tau-update} respectively.

To allow for incremental learning of our deep network, we exploit two additional components. The first is a memory which stores the most relevant samples of each class in $\mathcal{K}_t$. The relevance of a sample $(x,k)$ is determined by its distance $d_k(x)$ to the class mean $\mu_k$ \ie the lower is the distance, the higher is the relevance of the sample. The memory is used to augment the training set $\mathcal{T}_{t+1}$, allowing to update the mean estimates of the classes in $\mathcal{K}_t$ as the network is trained using samples of novel ones. In order to avoid an unbounded growth, the size of the memory is kept fixed and it is pruned after each incremental step to make room for instances of novel classes. The pruning is performed by removing, for each class in $\mathcal{K}_t$, the instances with lowest relevance}. 

The second component is a batch sampler 
which makes sure that, independently from the size of the memory, a given ratio of the batch is composed by samples taken from the memory. This allows to avoid 
biasing the incremental learning procedure towards novel categories, in the case their number of samples is much larger than the memory size. {Additionally, we add a distillation loss \cite{hinton2015distilling} which act as regularizer and 
 avoids the forgetting of previously learned features.} Denoting as $\phi_\Theta^{\mathcal{K}_{t-1}}$ the network trained on the set of known classes, the distillation loss is defined as:
\begin{equation}
\ell^{\text{distill}}(x_i) =||\phi_\Theta(x_i)- \phi_\Theta^{\mathcal{K}_{t-1}}(x)||
\end{equation}
The overall loss is thus defined as:
\begin{equation}
\label{eq:il-loss}
 \mathcal{L}=\frac{1}{|\mathcal{T}_t|}\sum_i \left( \ell^{\text{cl}}(x_i,k_i) + \lambda \ell^{\text{distill}}(x_i) \right)
\end{equation}
where $\lambda$ is an hyperparameter balancing the contribution of $\ell^{\text{distill}}$ within $\mathcal{L}$.


\subsubsection{Web-aided OWR}
\label{sec:web-download}
A restricting assumption of the traditional NNO algorithms in \cite{bendale2015towards,de2016online} is that data and labels of unknown classes are made available to the model by an oracle.
In practice, especially in robotics applications, this assumption is highly unrealistic since: i) the labels of samples of unknown categories are, by definition, unknown; 
ii) images of the unknown classes for incrementally updating the model are usually unavailable, since it is impossible to have a preloaded 
database containing all possible classes existing in the real world.
{A possible solution the aforementioned issues could involve the usage of information available on the Web. In 
 in this work we implement a first, very simple pipeline which tries to make use of the Web within the OWR procedure. } 
In particular, once an object is recognized as unknown, we query the Google Image Search engine\footnote{https://images.google.com/} to retrieve the closest keyword to the current image. Obviously the retrieved label might not be correct \eg due to low resolution of the image or a non canonical pose of the object. Here we tackle this issue through an additional human verification step, 
leaving the investigation of this problem to future works. As a final step, we use the retrieved keyword to automatically download images from the Web. These images represent new training data for the novel category which can be used to incrementally train the deep network. 

\vspace{-4pt}
\section{Experiments}
\vspace{-2pt}
\label{sec:experiments}
\vspace{-4pt}
In this section we show the results of our experimental evaluation. We first evaluate the performance of the proposed DeepNNO algorithm on two publicly available image recognition datasets. Then, we describe how the proposed OWR framework has been embedded into a robotic platform.

\vspace{-4pt}
\subsection{Experimental Setting}
\subsubsection{Datasets and Baselines}
We test the performances of our model on two datasets: CIFAR-100 \cite{krizhevsky2009learning} and Core50 \cite{lomonaco2017core50}. 
CIFAR-100 is a standard benchmark for testing the visual incremental learning  algorithms \cite{rebuffi2017icarl}. The dataset contains 100 different categories. In our experiments in the OWR setting, we split the dataset in two parts: 50 classes are considered as known categories, while the other 50 are the set of unknown classes. We consider 20 classes in the initial training set and we incrementally add the remaining known and unknown classes.

Core50 is a recently introduced benchmark for incremental learning methods which depicts images recorded in an egocentric setting. 
The dataset contains images of 50 objects corresponding to 10 semantic categories gathered under 11 different acquisition conditions. Following the standard protocol described in \cite{lomonaco2017core50}, we test on 3 sequences (sequences 3, 7, 10), using the remaining sequences for training. Since different sequences are taken under different acquisition conditions, this dataset represents a very challenging benchmark for object recognition. We split the dataset in two parts: 5 classes are considered known and the others belong to the unknown set.

Following \cite{bendale2015towards}, we evaluate the performances of our approach as classification accuracy and compare our method against its shallow counterpart, \ie the NNO algorithm. 
For each dataset we perform five different experiments, randomly selecting the classes in the known and unknown sets. The final performances are obtained by averaging results. 

\subsubsection{Networks architectures and training protocols}
Following \cite{he2016deep}, in our experiments we adopt the ResNet-18 architecture. For the experiments on CIFAR-100 we rescale the images to $32\times32$ pixels and perform random cropping and mirroring. 
We train the network from scratch on the initial set of known classes (batch-size 64, 120 epochs, learning rate 1.0, momentum 0.9, weight decay $10^{-5}$). Given this pretrained network, we apply both NNO and DeepNNO for learning an OWR recognition model. 
In the case of NNO we use the features extracted from the pretrained network to compute the class-specific mean vectors of novel categories, but we do not update the weight matrix $W$ and the threshold parameter $\tau$, as in \cite{bendale2015towards}. 
Differently, for DeepNNO we incrementally update the network parameters and train the network with SGD using the same hyperparameters adopted in the offline training stage, except for the number of epochs which is reduced to 40. We set $\lambda=1$, $w^+=1$ and $w^-=3$.  For sampling, we consider a fixed memory size of 2000 samples, 
constructing each batch by drawing 40\% of the instances from the memory. 
For experiments on Core50, we resize images to $128\times128$ pixels, employing the same network architecture and hyper-parameters of the CIFAR-100 experiments with the only difference that the number of epochs is reduced to 12 on the initial set of known classes and to 4 on the incremental training phase of DeepNNO.

We performed quantitative experiments in two different settings. In the first series of experiments we consider images of the two datasets to incrementally update our model. Differently, in a second set of experiments we consider images retrieved from the Web to train the OWR models. We consider images downloaded using keywords from three different engines: Google, Yahoo and Bing. Approximately 2000 images per semantic category have been retrieved. 

\vspace{-4pt}
\subsection{Experimental Results}
\vspace{-4pt}
\subsubsection{Analysis of the DeepNNO algorithm}
We first conduct an extensive experimental analysis of the proposed method considering the CIFAR-100 dataset. 

\begin{figure*}[t]
\minipage{0.3\textwidth}
   \includegraphics[width=1.0\textwidth,height=0.65\textwidth]{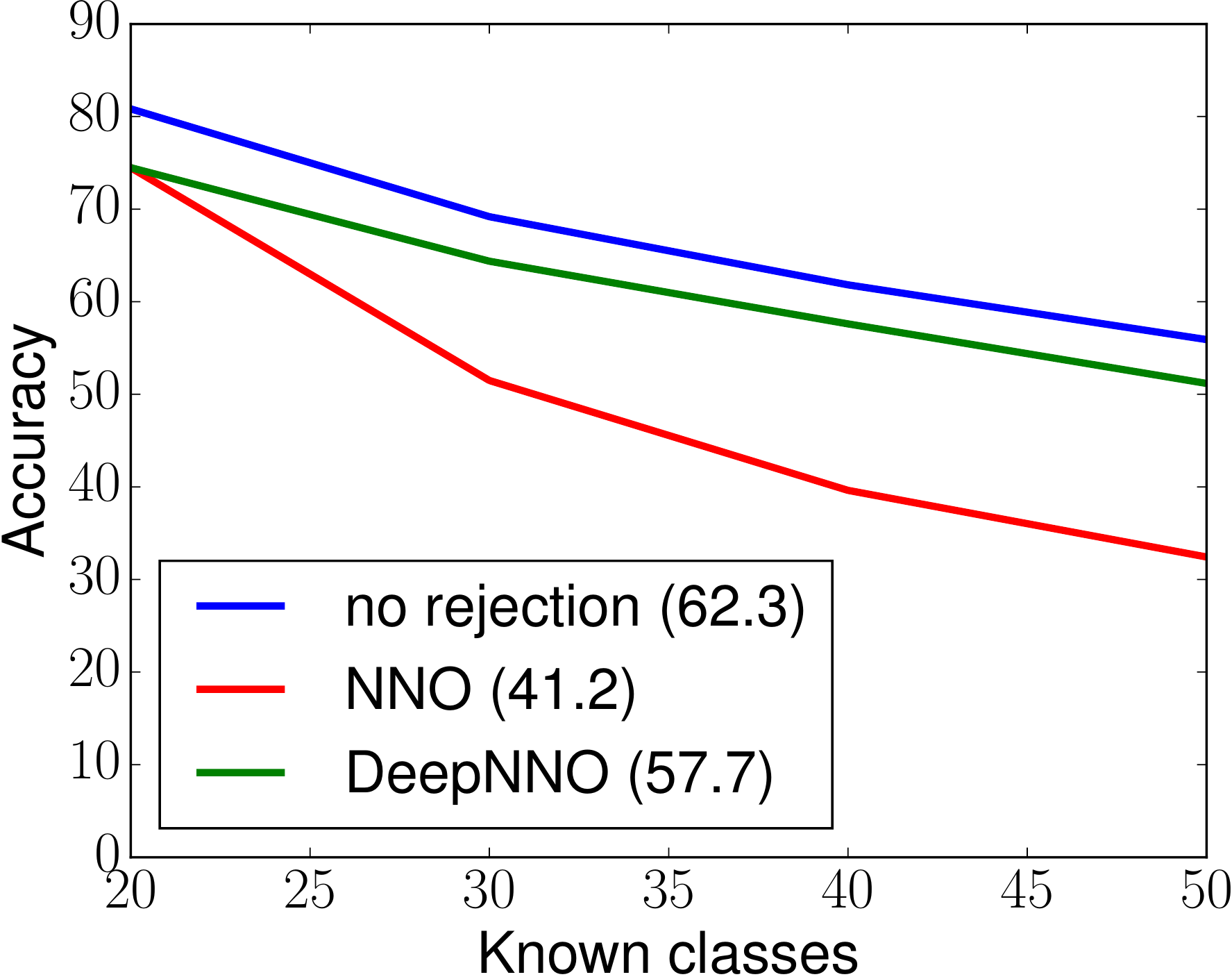}\vspace{-4pt}
   \caption{CIFAR-100 results in the closed world scenario.
  \vspace{-10pt}}
  \label{fig:res-cifar100-cw}
\endminipage\hfill
\minipage{0.3\textwidth}
   \includegraphics[width=0.95\textwidth,height=0.65\textwidth]{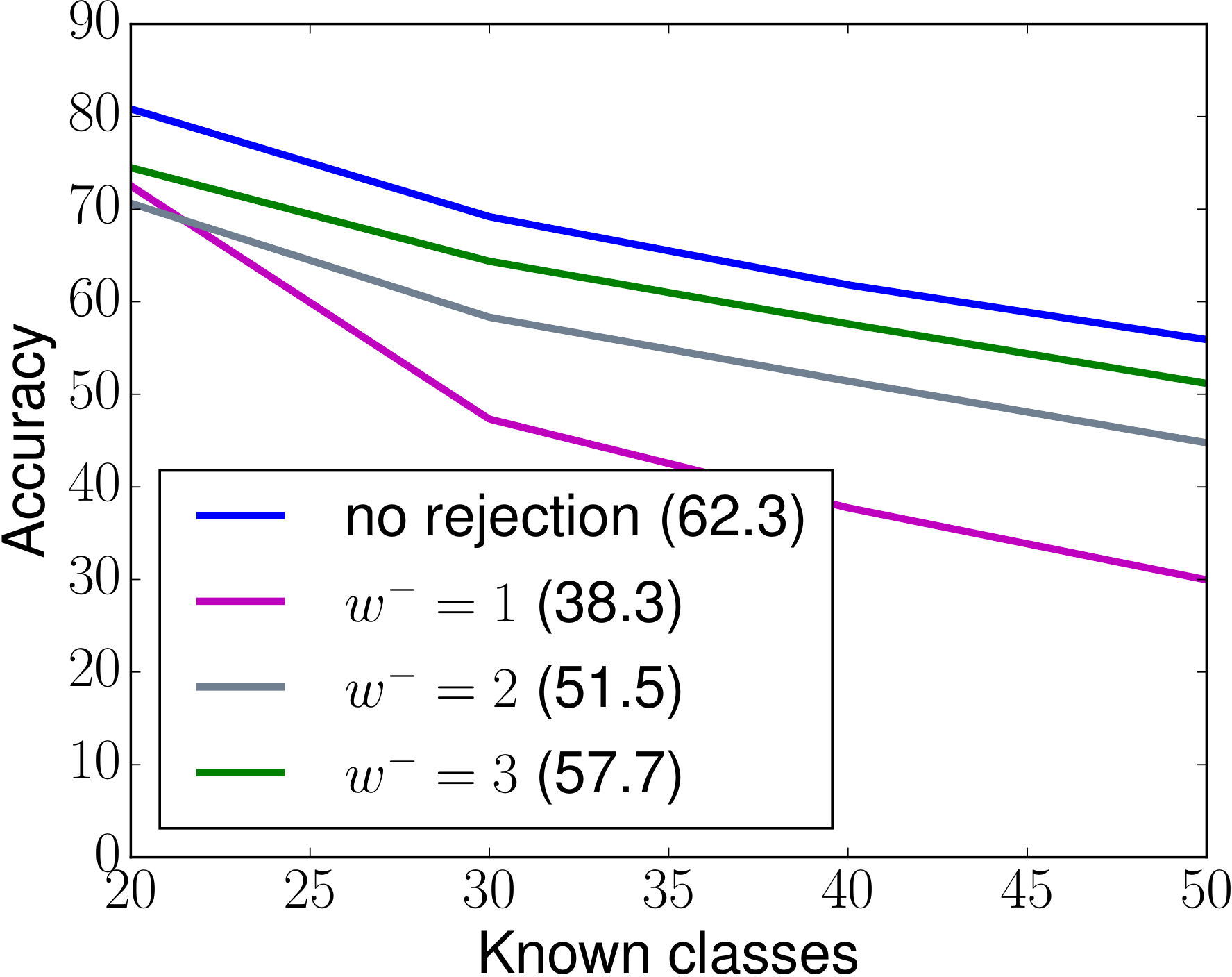}\vspace{-4pt}
   \caption{CIFAR-100 results of DeepNNO in the closed world scenario for different values of $w^-$.
  \vspace{-10pt}}
  \label{fig:ablation-margin}
\endminipage\hfill
\minipage{0.3\textwidth}%
    \includegraphics[width=0.95\textwidth,height=0.69\textwidth]{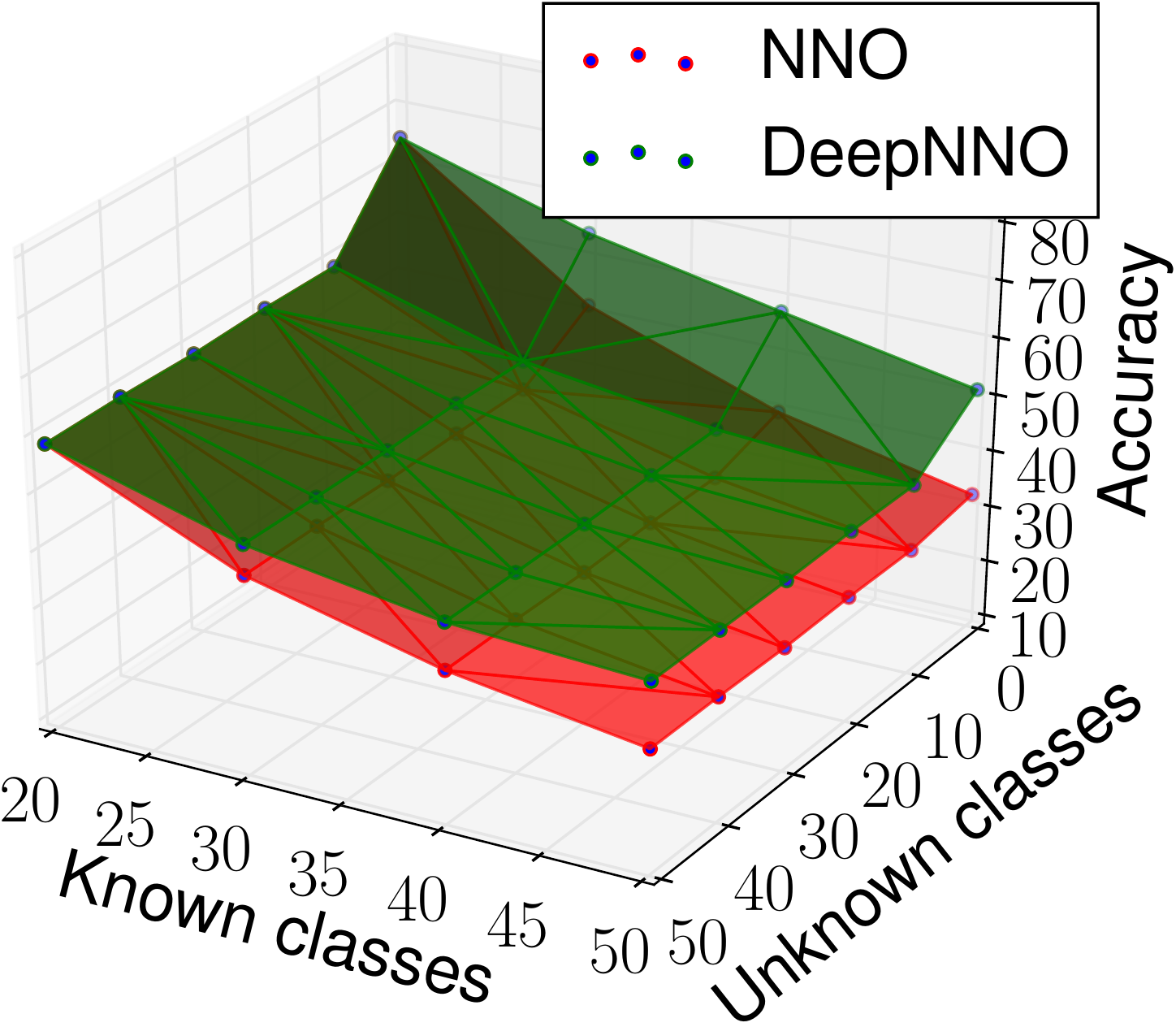}\vspace{-4pt}
   \caption{CIFAR-100: open world performances varying the number of known and unknown classes.
  \vspace{-10pt}}
  \label{fig:res-cifar100-ow3d}
   
\endminipage
\end{figure*}
\begin{figure*}[t]
\centering
\minipage{0.3\textwidth}
\centering
\includegraphics[width=0.95\textwidth,height=0.65\textwidth,height=0.65\textwidth]{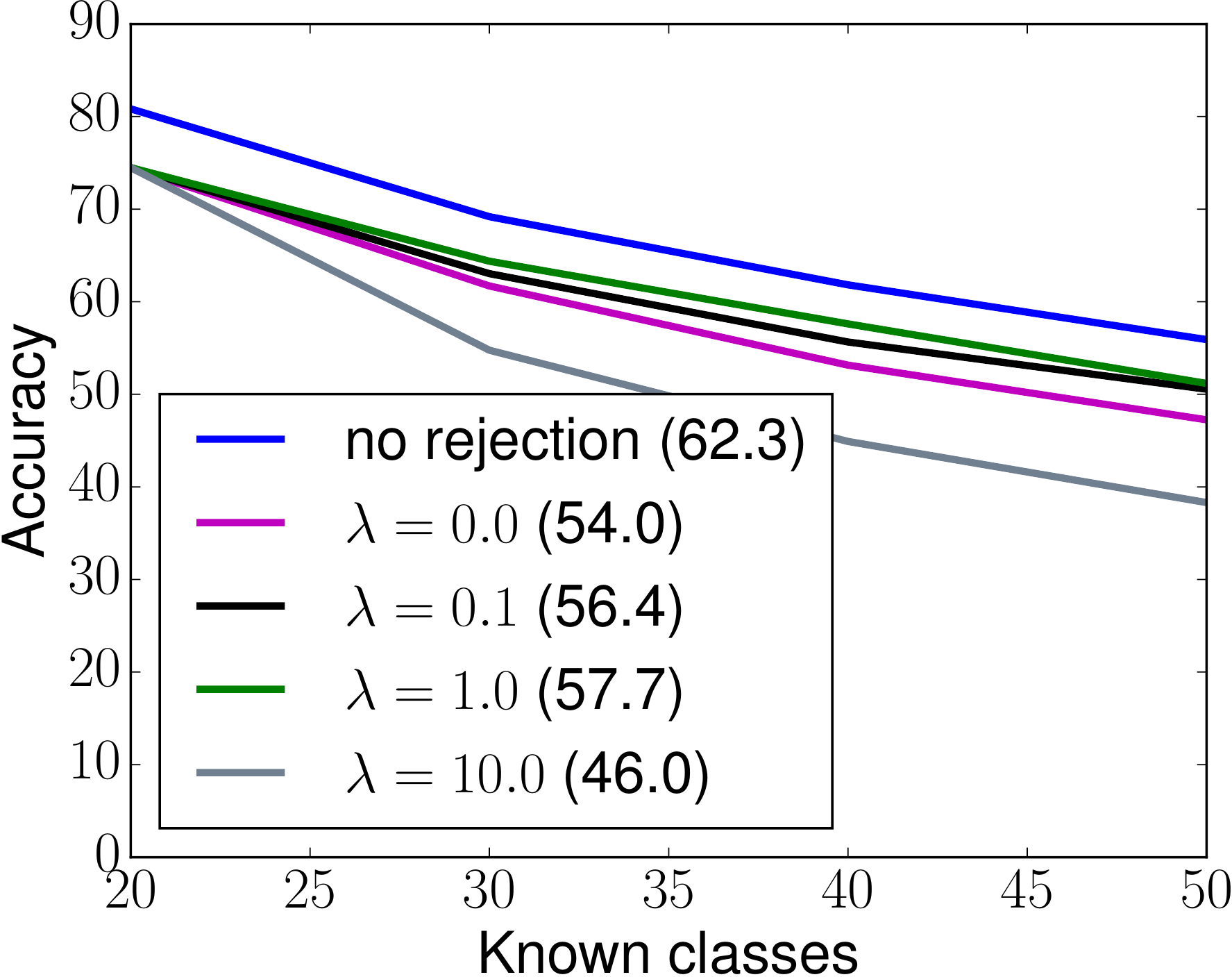}
   \caption{CIFAR-100 results of DeepNNO in the closed world scenario for different values of $\lambda$.
   \vspace{-14pt}}\vspace{-4pt}
   \label{fig:ablation-distill}
\endminipage\hfill
\minipage{0.3\textwidth}
\centering
    \includegraphics[width=0.95\textwidth,height=0.65\textwidth]{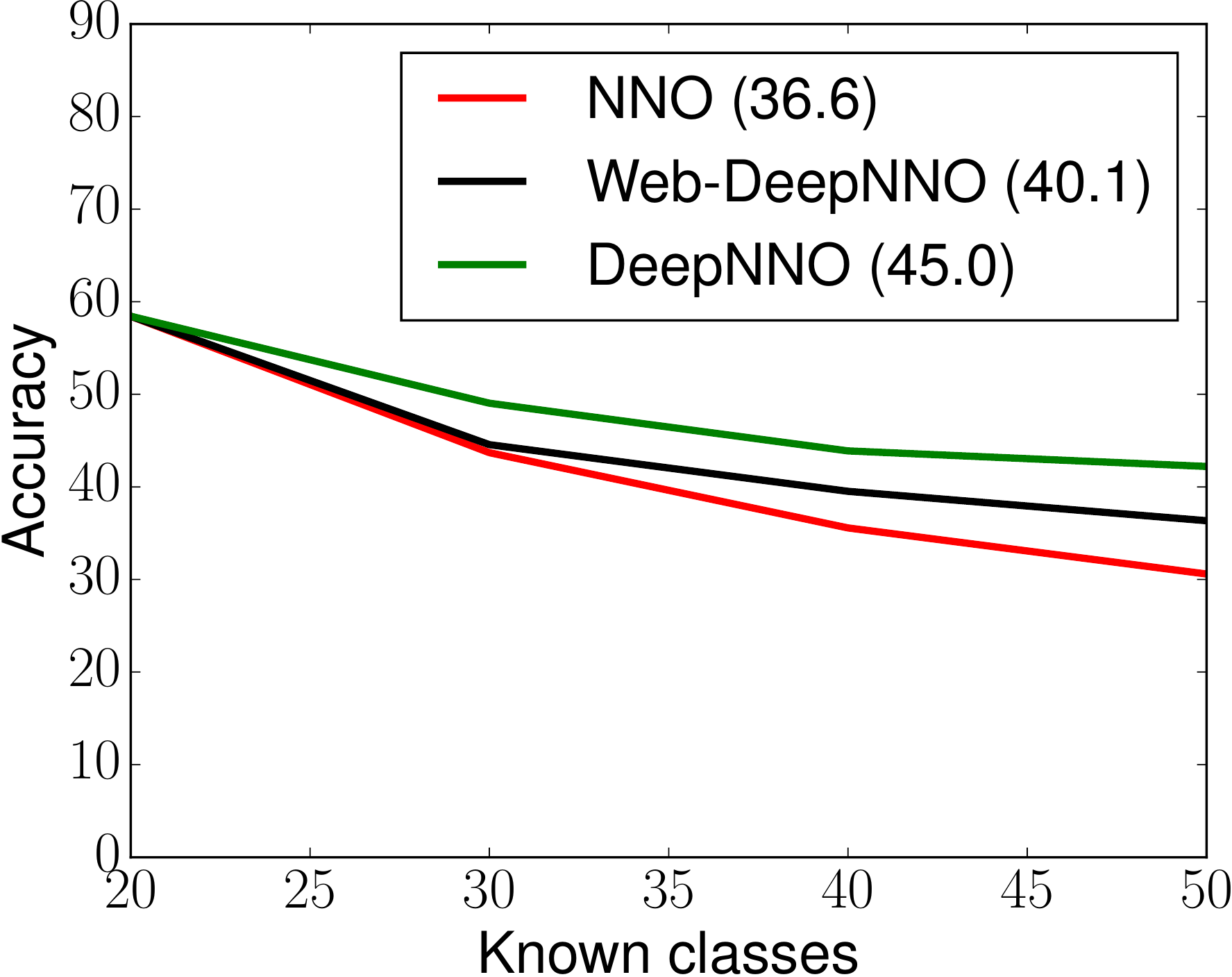}\vspace{-4pt}
      \caption{CIFAR-100: performances of Web-aided OWR in the open world scenario, with 50 unknown classes.
   \vspace{-14pt}}
   \label{fig:web-cifar100}
\endminipage\hfill
\minipage{0.3\textwidth}%
    \includegraphics[width=0.95\textwidth,height=0.65\textwidth]{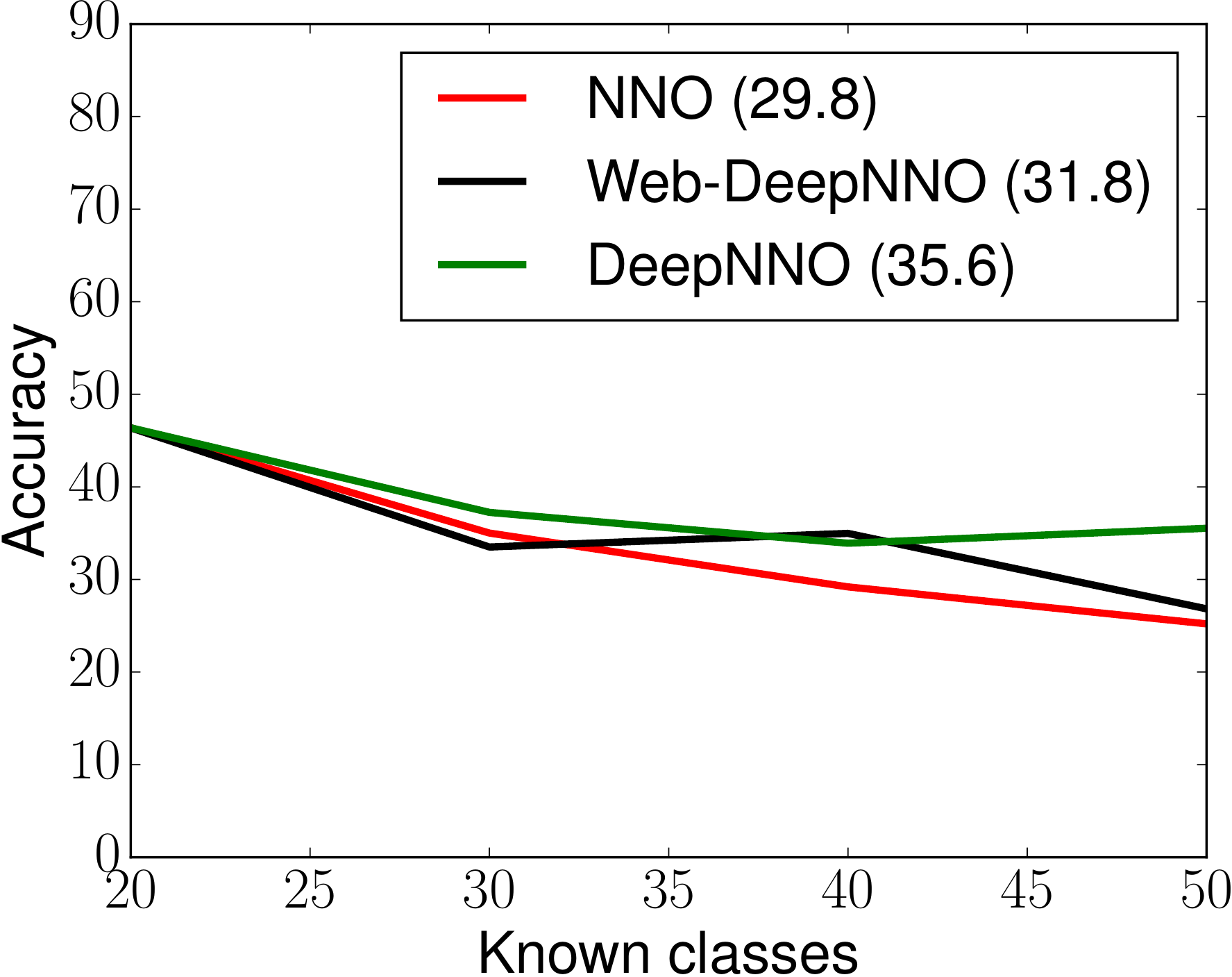}\vspace{-4pt}
      \caption{Core50 dataset: performances of Web-aided OWR in the open world scenario, with 5 unknown classes. 
   \vspace{-14pt}}
   \label{fig:web-core50}
\endminipage
\end{figure*}

In order to demonstrate the effectiveness of our classification algorithm, we start by performing experiments in the closed world scenario, \ie measuring the performances considering only the set of known classes.  
We compare the performance of DeepNNO with NNO and DeepNNO without rejection option (\ie \textit{DeepNNO-no rejection}). The latter baseline method is our upper bound in terms of performances in the closed world, since it does not reject any instance of known classes (\ie it does not identify samples of known classes as unknowns). This baseline is used to demonstrate the validity of the proposed method for setting the threshold $\theta$. The results are shown in Fig. \ref{fig:res-cifar100-cw} where the numbers between parenthesis denote the average accuracy among the different incremental steps. From Fig. \ref{fig:res-cifar100-cw} it is possible to draw two observations. First, there is a large gap between the performances of DeepNNO and NNO, with our model outperforming its non-deep counterpart by more than $16\%$ on average and by more than $20\%$ after all the incremental steps. The improved performance of our method can be ascribed to the fact that, by dynamically updating the learned feature representations, DeepNNO is able to better adapt the learned classifier to novel semantic concepts. Second, DeepNNO achieves results close to DeepNNO without rejection. This indicates that, thanks to the proposed approach for setting the threshold $\theta$, our method only rarely identifies samples of known classes as belonging to an unknown category. We believe this is mainly due to the introduction of the different weighting factors $w^-$ and $w^+$ while updating $\theta$. 
This observation is confirmed by results shown in Fig.~\ref{fig:ablation-margin} 
 which analyzes the effect of varying $w^-$ with $w^+$ fixed to 1. As $w^-$ decreases, the accuracy decreases as well, due to the higher value reached by $\theta$ which leads to wrongly reject many samples, classified as instances of unknown classes.
%
%
%

As a second experiment, we compare the performances of DeepNNO and NNO in the open world recognition scenario varying the number of known and unknown classes. The results are shown in Fig. \ref{fig:res-cifar100-ow3d}, from which it is easy to see that DeepNNO outperforms its non-deep counterpart by a large margin. 
In fact, in this scenario, our model achieves an accuracy 9\% higher then standard NNO on average considering 50 unknown classes.
Moreover, this margin increases during the training: after all the incremental steps our model outperforms NNO by a margin close to 15\%.
It is worth noting that the advantages of our model are independent on the number of unknown classes, since DeepNNO constantly outperforms NNO in all settings.  





Another important component of our method is the distillation loss. This loss guarantees the right balance between learning novel concepts and preserving old features.
To analyze its impact, in Fig.~\ref{fig:ablation-distill} we report the performances of DeepNNO in the closed world scenario for different values of $\lambda$. From the figure it is clear that, without the regularization effect of the distillation loss, the accuracy significantly drops. 
On the other hand, a high value of $\lambda$ leads to poor performances and low confidence on the novel categories. Properly balancing the contribution of classification and distillation loss the best performance can be achieved. 




\subsubsection{Web-aided OWR}
In a second series of experiments we analyze the performance of the proposed OWR framework which exploits Web images in the incremental learning phase.
The results of our experiments are shown in Fig.~\ref{fig:web-cifar100} for CIFAR-100 and in Fig.~\ref{fig:web-core50} for Core50. As expected, considering images from the Web instead of images from the datasets lead to a decrease in terms of performance. However, the accuracy of Web-aided DeepNNO is still good, especially when compared with its non-deep counterpart. 

On the CIFAR-100 experiments we achieve a remarkable performance, with Web DeepNNO outperforming NNO by 3.5\% on average and by more than 5\% after all the incremental steps. We highlight that these results have been achieved exploiting only noisy and weakly labeled Web images. 
On the Core50 experiments, both DeepNNO and its Web-based version, achieve higher accuracy w.r.t. NNO. 
However, in this case the performance improvement with respect to NNO is more modest. We ascribe this behavior to the fact that there is a large appearance gap between Core50 images gathered in an egocentric setting and Web images. 
We believe that this issue can be addressed in future works by imposing some constraints on the quality of downloaded images and by coupling DeepNNO with domain adaptation techniques~\cite{patel2015visual,carlucci2017autodial,mancini2018kitting,mancini2018boosting} in order to reduce the domain shift between downloaded images and training data. 

 
\begin{figure}[t]
    \centering
    \includegraphics[width=1.\columnwidth, height=0.45\columnwidth]{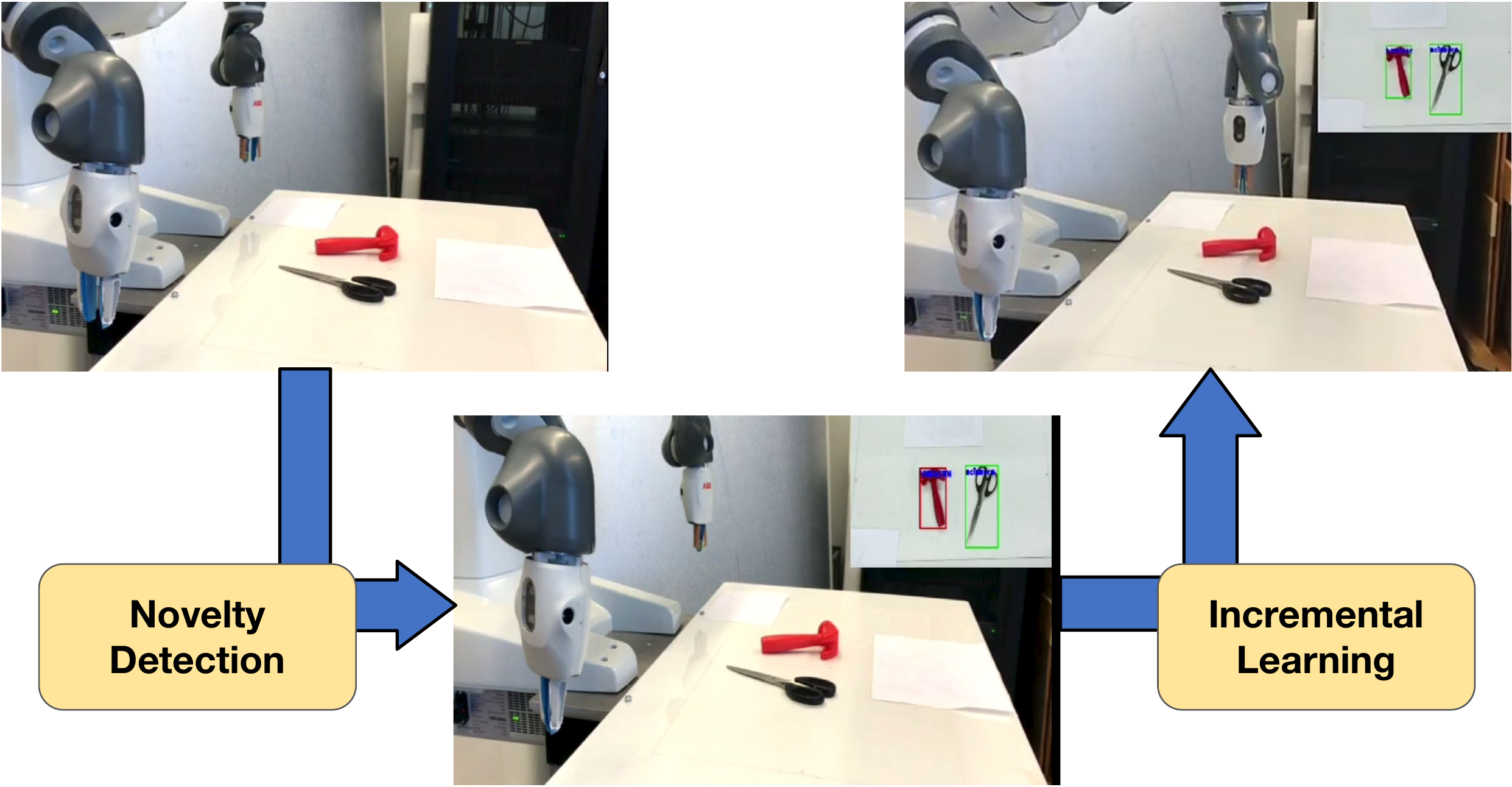}
    \caption{{Qualitative results of deployment on a robotic platform. The robot recognizes an object as unknown (\ie the red hammer, bottom) and adds it to the knowledge base through the incremental learning procedure (top right).}
     \vspace{-20pt}}
    \label{fig:task}
  \end{figure}

\subsubsection{Deployment on a Robotic Platform}
Finally, we tested the proposed Web DeepNNO algorithm by integrating it into a visual object detection framework and running it on a Yumi 2-arm manipulator equipped with a Kinect. We have used the Faster-RCNN framework in~\cite{ren2015faster} with the ResNet-101 architecture~\cite{he2016identity} as backbone. We pretrained the network on the COCO dataset \cite{lin2014microsoft}, 
after replacing the standard fully-connected classifier with the proposed DeepNNO. We performed an open world detection experiment by placing multiple objects (known and unknowns) in the workspace of the robot. Whenever a novel object is detected, the robot tries to get the corresponding label from Google Image Search, with using the cropped image of the unknown object. In case the label is not correct, a human operator cooperates with the robot and provides the right label. The provided label is used by the robot to automatically download the images associated to the novel class from the Web sources. These images are then used to update the classification model. 

{Figure~\ref{fig:task} shows a qualitative result associated to our experiment. For instance, the robot is able to correctly detect the red hammer as unknown. The full example is available in the supplementary material.}

\vspace{-4pt}
\section{Conclusions}
\label{sec:conclusions}
\vspace{-4pt}
This paper addresses the problem of continuous learning of visual objects for a robot system challenged with knowledge gaps. We presented a principled algorithm rooted into the Open World Recognition framework, that couples the flexibility of non-parametric learning methods with the power of end-to-end deep learning. to further overcome the need for annotated images for the new detected classes, we proposed a simple strategy towards mining annotated data from the Web. Experiments on two different databases and on a mobile robot platform show the promise of our approach. Future work will further investigate webly supervised approaches with the goal of pushing the envelope in life-long learning of autonomous systems.

\bibliographystyle{IEEEtran}
\bibliography{IEEEabrv,root}

\end{document}